\documentclass[10pt,twocolumn,letterpaper]{article}

\usepackage[pagenumbers]{iccv} 

\usepackage{colortbl} 
\usepackage{amssymb}
\usepackage{xcolor}   
\usepackage{graphicx} 
\usepackage{xcolor}
\usepackage{multirow}

\definecolor{iccvblue}{rgb}{0.21,0.49,0.74}
\usepackage[pagebackref,breaklinks,colorlinks,allcolors=iccvblue]{hyperref}

\def\paperID{1666} 
\def\confName{ICCV}
\def\confYear{2025}

\title{Global-Aware Monocular Semantic Scene Completion with State Space Models}

\author{
    Shijie Li$^1$\quad Zhongyao Cheng$^1$\quad  Rong Li$^2$ \quad Shuai Li$^3$ \\[1ex]
    \quad Juergen Gall$^3$\quad Xun Xu$^1$\quad Xulei Yang$^1$
    \\[1ex]
    $1.$ I$^2$R, A*STAR~~~
    $2.$ AI Thrust, HKUST(GZ)~~~
    $3.$ Bonn University
}

\begin{document}
\maketitle
\begin{abstract}
Monocular Semantic Scene Completion (MonoSSC) reconstructs and interprets 3D environments from a single image, enabling diverse real-world applications. However, existing methods are often constrained by the local receptive field of Convolutional Neural Networks (CNNs), making it challenging to handle the non-uniform distribution of projected points (Fig. \ref{fig:perspective}) and effectively reconstruct missing information caused by the 3D-to-2D projection.
In this work, we introduce GA-MonoSSC, a hybrid architecture for MonoSSC that effectively captures global context in both the 2D image domain and 3D space. Specifically, we propose a Dual-Head Multi-Modality Encoder, which leverages a Transformer architecture to capture spatial relationships across all features in the 2D image domain, enabling more comprehensive 2D feature extraction.
Additionally, we introduce the Frustum Mamba Decoder, built on the State Space Model (SSM), to efficiently capture long-range dependencies in 3D space.
Furthermore, we propose a frustum reordering strategy within the Frustum Mamba Decoder to mitigate feature discontinuities in the reordered voxel sequence, ensuring better alignment with the scan mechanism of the State Space Model (SSM) for improved 3D representation learning. We conduct extensive experiments on the widely used Occ-ScanNet and NYUv2 datasets, demonstrating that our proposed method achieves state-of-the-art performance, validating its effectiveness. The code will be released upon acceptance.
\end{abstract}    
\section{Introduction}
\label{sec:intro}

Semantic Scene Completion (SSC) plays a crucial role in various 3D Vision related applications, such as robotic navigation \cite{desouza2002vision, tremblay2020indirect}, augmented reality \cite{azuma1997survey}, and autonomous driving \cite{yurtsever2020survey, jiang2024symphonize, huang2023tri}. Its goal is to complete incomplete 3D scenes, typically represented as a voxel grid, while simultaneously detecting the relevant semantic information.

\begin{figure}[t]
    \centering
    \includegraphics[width=\linewidth]{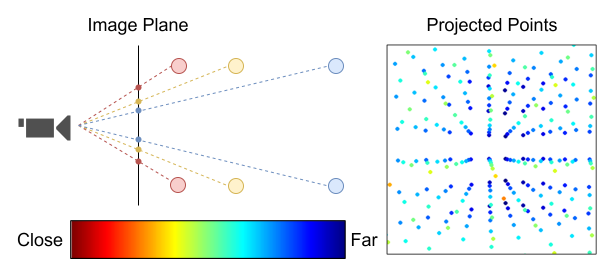}
    \caption{Perspective projection results in an imbalanced distribution of projected points, where nearby points are more widely scattered while distant points are more densely packed. The real projection map is shown on the right.}
    \vspace{-6mm}
    \label{fig:perspective}
\end{figure}

This work focuses on Monocular Semantic Scene Completion (MonoSSC) specifically tailored for indoor environments. Many recent SSC studies \cite{tian2024occ3d, li2023fb, wang2021learning} focus on autonomous driving scenarios, aiming to complete incomplete 3D scenes using multi-view images or 3D LiDAR point clouds as input. However, such rich sensory data is often unavailable in indoor environments where usually only a monocular camera is available.
Thus MonoSSC is proposed, which can infer 3D geometry and semantic information from a single image. Compared to SSC in autonomous driving scenarios, indoor MonoSSC is typically more challenging due to  less informative sensory data and increased complexity of indoor environments with intricate structures.

The capability of current MonoSSC solutions is often constrained by the inherent limitations of CNN architectures. These methods typically start by extracting 2D features using 2D Convolutional Neural Networks (2D CNNs). Each 3D voxel is then projected onto the 2D feature map to retrieve the corresponding features. Subsequently, 3D Convolutional Neural Networks (3D CNNs) refine these 3D voxel features and generate the final prediction. As shown in Fig. \ref{fig:perspective}, the projected 3D voxel centers are unevenly distributed on the 2D image plane due to perspective projection. Specifically, the projections of 3D points closer to the camera are more widely scattered, while those of farther points are more densely packed.
CNN architectures adopt a fixed-weight encoding strategy. The resulting feature map usually cannot capture this variance or adapt to the non-uniform distribution, leading to ineffective 2D feature representation and suboptimal spatial reasoning in the MonoSSC task.
Furthermore, the 3D-to-2D projection inherently results in information loss along the depth dimension. To maximally recover the lost information, the model needs to capture global dependencies in 3D space. However, traditional CNN architectures are restricted by their local receptive fields, making it challenging to effectively capture long-range spatial relationships in 3D space.

To address these challenges, in this work, we propose a MonoSSC method named \textbf{GA-MonoSSC}, which possesses global awareness in both 2D image space and 3D space. Specifically, we propose a Dual-Head Multi-Modality Encoder ($\mathbf{DM_{enc}}$), which leverages a Transformer architecture to capture spatial relationships across 2D features.This enables interactions among imbalanced projected points, resulting in a more informative and robust representation. Additionally, multi-modality information, including semantic and geometric features, is learned separately, allowing the model to capture more nuanced representations compared to previous methods that process them jointly.

\begin{figure}[t]
    \centering
    \includegraphics[width=\linewidth]{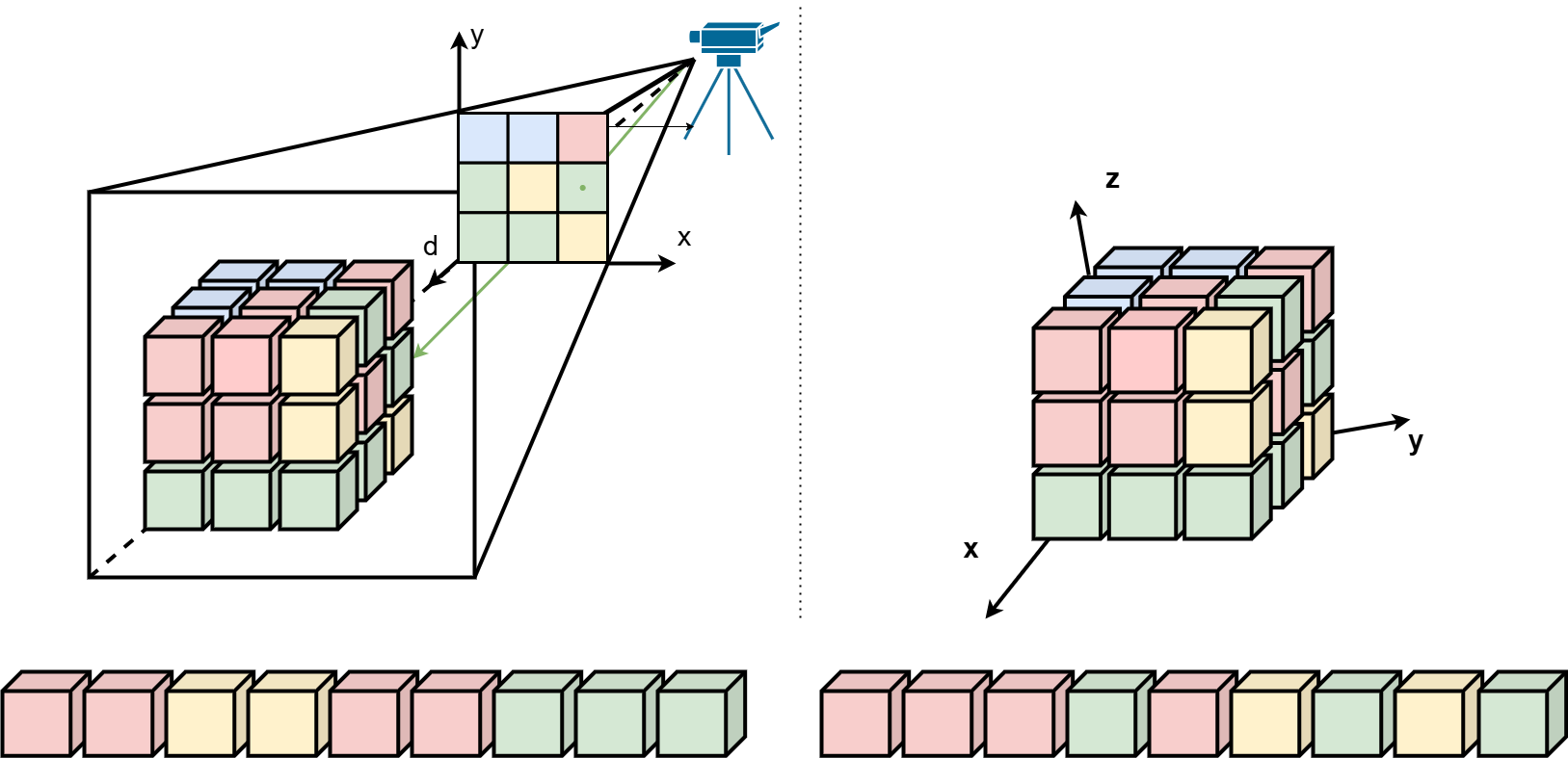}
    \caption{Arranging voxel features into a sequence based on their 3D coordinates introduces discontinuities (right). This issue can be mitigated by sorting voxels within the frustum space, where voxel features are distributed across the image plane and exhibit strong similarity along the depth dimension (left).}
    \vspace{-6mm}
    \label{fig:discontinued}
\end{figure}

The 3D model is also expected to exhibit global awareness to effectively reconstruct missing information in 3D space. However, Transformer architectures are impractical for this task due to their quadratic computational complexity, which becomes prohibitive given the typically large input sequence length in 3D space. Instead, we adopt the Selective State Space Model (SSM) \cite{gu2023mamba}, specifically Mamba, which efficiently models long-range dependencies with linear complexity.
Mamba typically arranges voxel features as sequential data based on their 3D spatial coordinates and processes them accordingly. However, we observe that this approach introduces discontinuities in the reorganized sequential data due to the 2D-to-3D unprojection process, as illustrated in Fig. \ref{fig:discontinued}. To address this issue, we propose the Frustum Mamba Decoder ($\mathbf{FM_{Dec}}$), which incorporates a Frustum Mamba Layer specifically designed to reorder voxels in frustum space.
This ensures a more structured and continuous feature flow, better aligning with the underlying data distribution.

Our main contributions can be summarized as follows.

\begin{itemize}
    \item We propose a hybrid framework, GA-MonoSSC, that integrates Transformer and Mamba architectures to achieve global awareness in both the 2D image domain and 3D space. The proposed method achieves state-of-the-art performance on the Occ-ScanNet and NYUv2 datasets.
    \item We design a Dual-Head Multi-Modality Encoder that employs a Transformer to capture spatial relationships across 2D features, mitigating the impact of the non-uniform distribution of projected points. Additionally, semantic and geometric features are learned separately, enabling the model to capture finer details and enhance representation quality.
    \item We integrate Mamba \cite{gu2023mamba} to capture long-range dependencies in 3D space, enabling maximal recovery of lost information. We also introduce the Frustum Mamba Layer, which scans voxels in frustum space, addressing the feature discontinuities caused by 2D-to-3D unprojection.
\end{itemize}

\section{Related Works}
\label{sec:related_work}

\subsection{Monocular Semantic Scene Completion}
Monocular Semantic Scene Completion (SSC) aims to reconstruct a complete 3D structure with semantic labels from a single 2D image. This monocular approach was introduced by MonoScene \cite{cao2022monoscene}, advancing beyond earlier SSC methods \cite{li2020anisotropic, liu2018see, song2017semantic, zhang2018efficient, zhong2020semantic, cai2021semantic, tang2022not, zhang2019cascaded,jiang2024symphonize,zhang2023occformer,yu2024context,wang2024occrwkv,yu2024language,liang2025skip} by relying solely on 2D visual input, thus removing the need for additional 3D data sources. A key limitation in MonoScene \cite{cao2022monoscene} is the ambiguity in the unprojection operation, which maps 2D features into 3D space but struggles with depth accuracy. To address this, recent approaches have focused on refining depth information within the unprojection process. NDC-Scene \cite{yao2023ndc} introduced a Depth-Adaptive Dual Decoder that upscales and merges 2D and 3D feature maps, enhancing spatial accuracy. Similarly, ISO \cite{yu2024monocular} proposes directly predicting depth values and incorporating them into the unprojection operation, leading to improved performance. However, these solutions are constrained by limited global information modeling capacity in 3D space due to efficiency issues. In this work, we leverage the Mamba  to capture global information efficiently, overcoming these limitations.

\subsection{State Space Models}

State Space Models (SSMs), including Mamba \cite{gu2023mamba}, S4 \cite{gu2021efficiently}, and S4nd \cite{nguyen2022s4nd}, have gained considerable attention due to their efficiency in handling long-range dependencies and large-scale data with linear time complexity. Mamba, in particular, integrates selective mechanisms that enhance this capability, making it especially notable. These innovations have led to successful adaptations in computer vision \cite{hatamizadeh2024mambavision,rahman2024mamba,patro2024simba,yang2024vivim}.  VMamba \cite{zhu2024vision} incorporates a cross-scan module and Vision Mamba \cite{zhu2024vision} utilizes a bidirectional SSM architecture, both optimized for visual data processing.
The architecture has also been extended to process 3D data, such as point clouds, where approaches like \cite{liang2024pointmamba} use a 3D Hilbert curve to rearrange input patches, and \cite{liu2024point} employs an octree-based ordering to efficiently capture spatial relationships. Building on these advancements, we propose a Mamba-based network specifically optimized for processing 3D voxel data, leveraging the strengths of SSMs to handle complex spatial data more effectively and efficiently. 


\section{Methodology}
\label{sec:method}

In this section, we present a detailed description of the proposed \textbf{GA-MonoSSC}. We introduce the State Space Models and the Mamba Module as essential preliminaries in Sec. \ref{sec:pre}. Next, we provide an overview of the overall architecture in Sec \ref{sec:overview} and delve into the specifics of the main components in Sec. \ref{sec:2d_bb} and Sec. \ref{sec:3d_bb}. The employed supervision is shown in Sec. \ref{sec:supervision}.

\subsection{Preliminaries}
\label{sec:pre}

\textbf{State Space Models.} 
To set the stage for the Mamba module, we first revisit the concept of state space models (SSMs). Rooted in control theory, SSMs operate by mapping inputs to outputs via an intermediary hidden state, making them well-suited for processing sequential data. For discrete inputs, SSMs can be formally represented as follows:
\begin{equation}
    h_k = \overline{A} h_{k-1} + \overline{B} x_k,
\end{equation}
\begin{equation}
    y_k = \overline{C} h_k,
\end{equation}

where $ k $ is the sequence number, $ \overline{A} $, $ \overline{B} $, $ \overline{C} $ are matrices that represent the discretized parameters of the model, which involve the sampling step $ \Delta $. The $ x_k $, $ y_k $, and $ h_k $ representing the input, output, and hidden state of the system, respectively. Structured State Space Sequence models (S4) \cite{gu2021efficiently} introduce a novel parameterization for SSMs, enabling more efficient computation.

\textbf{Mamba Module.} Mamba introduces an adaptation to S4 models \cite{gu2021efficiently}, making the matrices $ \overline{B} $ and $ \overline{C} $, as well as the sampling size $ \Delta $, dependent on the input, with this dependency arising from incorporating the sequence length and batch size of the input. This enables dynamic adjustment of the matrices for each input token, allowing Mamba to $ \overline{B} $ and $ \overline{C} $ to dynamically adjust state transitions based on the input, thereby enhancing the model’s content awareness.

\subsection{Overview}
\label{sec:overview}

\begin{figure*}
    \centering
    \includegraphics[width=\linewidth]{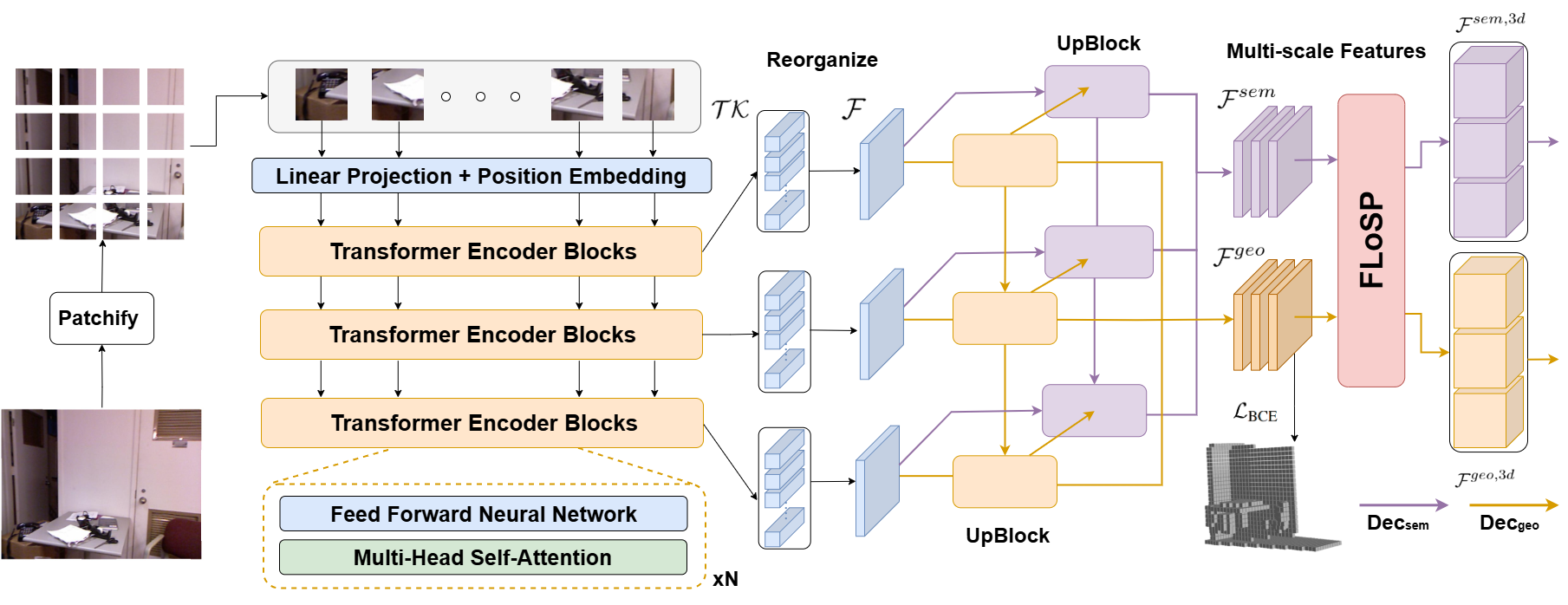}
    \caption{The overview of the proposed Dual-head Multi-modality Encoder (DM$_{\textbf{Enc}}$) which will extract multi-scale geometry and semantic features separately}
    \vspace{-5mm}
    \label{fig:arch2d}
\end{figure*}

\begin{figure*}[h]
    \centering
    \includegraphics[width=\linewidth]{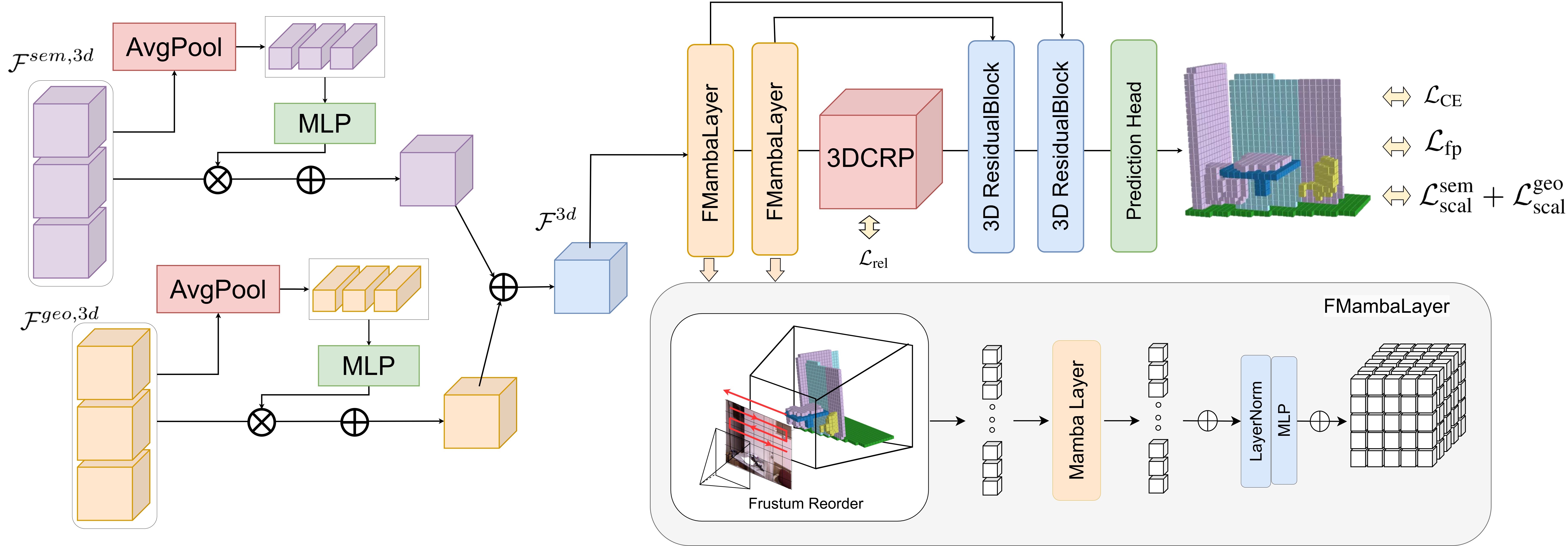}
    \caption{Architecture of the proposed Frustum Mamba Decoder $\mathbf{FM_{dec}}$. $\mathbf{FM_{dec}}$ adopts an encoder-decoder architecture, with the proposed Frustum Mamba Layer integrated into each stage of the encoder to capture long-range dependency. The Frustum Reordering Strategy is introduced to arrange voxels into a sequence within frustum space rather than the original 3D space, effectively mitigating feature discontinuities and improving sequential consistency.}
    \vspace{-5mm}
    \label{fig:arch3d}
\end{figure*}

\textbf{GA-MonoSSC} takes a monocular image $\mathcal{X}^{2d} \in \mathcal{R}^{C \times H^{2d} \times W^{2d}}$ as input and outputs a 3D semantic occupancy map $\mathcal{X}^{3d} \in \mathcal{R}^{N_c \times H^{3d} \times W^{3d} \times L^{3d}}$, where $N_c$ is the number of total semantic classes:
\begin{equation}
    \mathcal{X}^{3d} = \textbf{GA-MonoSSC}(\mathcal{X}^{2d})
\end{equation}
The input monocular image is first processed by a Dual-head Multi-modality encoder ($\mathbf{DM_{enc}}$), shown in Fig. \ref{fig:arch2d}, which separately extracts global-aware multi-scale semantic and geometric features. These features are then unprojected into 3D voxel space using Feature Line of Sight Projection ($\mathbf{FLoSP}$) \cite{cao2022monoscene}. In 3D space, multi-scale geometry and semantic features are first fused as a unified representation, which is then processed by the proposed Frustum Mamba Decoder ($\mathbf{FM_{dec}}$), shown in Fig. \ref{fig:arch3d}, to capture long-range dependencies effectively.
The 3D predictions will be decoded through a prediction head finally.

\subsection{Dual-head Multi-modality Encoder (DM$_{\textbf{Enc}}$)}
\label{sec:2d_bb}

We first introduce the Dual-Head Multi-Modality Encoder ($\mathbf{DM_{enc}}$), as shown in Fig. \ref{fig:arch2d}, which utilizes a Transformer architecture to extract multi-scale geometric and semantic features while effectively modeling spatial relationships among 2D features.
In previous solutions, CNN architectures have been widely used for 2D feature extraction. However, their fixed-weight encoding strategy and limited local receptive field hinder their ability to capture the non-uniform distribution of projected points. This limitation results in ineffective feature representation and suboptimal spatial reasoning for the MonoSSC task, as illustrated in Fig. \ref{fig:perspective}.
To address this challenge, the model requires the ability to capture long-range dependencies, making the Transformer architecture an ideal choice.

$\mathbf{DM_{enc}}$ consists of a transformer-based encoder, $\mathbf{Enc}$, which comprises $N_l$ transformer encoder blocks and outputs multi-scale 2D tokens. These tokens are then fed into two modality-specific decoders, $\mathbf{Dec_{geo}}$ and $\mathbf{Dec_{sem}}$, to decode geometric and semantic information.
$\mathbf{Enc}$ receives input single-view image $\mathcal{X}^{2d}$ and output multi-scale tokens:
\begin{equation}
    \mathcal{TK}= \mathbf{Enc}(\mathcal{X}^{2d})
\end{equation}
$\mathcal{TK} = \{ TK_l \in \mathbb{R}^{N_{tk} \times C} \mid l \in (1, N_l) \}$ represents the set of tokens obtained after each transformer encoder block $l$. $N_{tk}$ is the number of tokens. To recover spatial information, the tokens $TK_l$ at each scale are reorganized into a spatial feature map:
\begin{equation}
    \mathcal{F}_l = \mathbf{Reorganize}(\mathcal{TK}_l)
\end{equation}
The output feature maps $\mathcal{F} = \{ \mathcal{F}_l \in R^{C \times h \times w} | l \in (1, N_l)\}$ where $h$ and $w$ are spatial shape after reorganization, will be fed into the decoder $\mathbf{Dec_{geo}}$ to decode multi-scale geometric features:
\begin{equation}
    \mathcal{F}^{geo} = \mathbf{Dec}_{geo}(\mathcal{F})
\end{equation}
where $\mathcal{F}^{geo} = \{ \mathcal{F}^{geo}_l \in R^{C \times H^{2d}_l \times W^{2d}_l} | l \in (1, N_l)\}$. Specially, $\mathbf{Dec_{geo}}$ contains several $\mathbf{UpBlock}$ each can be formulated as:
\begin{equation}
    \mathcal{F}^{geo}_l = \mathbf{Up_{2d}}(\mathcal{F}_l)
\end{equation}
\begin{equation}
    \hat{\mathcal{F}}_{l-1}^{geo} = \mathbf{Up_{2d}}(\mathbf{Conv_{2d}}(\mathcal{F}^{geo}_{l-1}))
\end{equation}
\begin{equation}
    \mathcal{F}^{geo}_l = \mathbf{Conv_{2d}}(\mathbf{BN}(\mathbf{ReLU}(\mathcal{F}^{geo}_l || \hat{\mathcal{F}}^{geo}_{l-1})))
\end{equation}
where $\mathbf{Up_{2d}}$ denotes upsample operation, $\mathbf{Conv_{2d}}$, $\mathbf{BN}$ and $\mathbf{ReLU}$ means convolution layer, batch normalization, and Rectified Linear Unit separately. $||$ denotes the concatenation operation.

The multi-scale semantic features $\mathcal{F}^{sem} = \{ \mathcal{F}^{sem}_l \in R^{C \times H^{2d}_l \times W^{2d}_l} | l \in (1, N_l)\}$ are derived similarly with decoder $\mathbf{Dec_{sem}}$ which share the same architecture to  $\mathbf{Dec_{geo}}$. One thing needs to be noticed is that the geometric features $\mathcal{F}^{geo}$ will be fed in to introduce more structure information:
\begin{equation}
    \mathcal{F}^{sem} = \mathbf{Dec}_{sem}(\mathcal{F}^{2d}, \mathcal{F}^{geo})
\end{equation}
Specially, at each scale $l$, the geometry feature map will be fused with an elementwise add:
\begin{equation}
   \mathcal{F}_l^{sem}  = \mathcal{F}_l^{sem} + \mathcal{F}_l^{geo}
\end{equation}
Due to the ambiguity introduced by information loss in the 3D-to-2D projection, we enforce 3D supervision on the output of $\mathbf{Dec_{geo}}$ to ensure that the extracted geometric features retain awareness of the 3D data distribution. Specifically, we achieve this by predicting depth from \(\mathbf{Dec}_{geo}\), which is subsequently converted into a 3D occupancy representation, where the supervision is applied.

\subsection{Frustum Mamba Decoder (FM$_{\textbf{Dec}}$)}
\label{sec:3d_bb}

The extracted multi-scale, geometry and semantic features, $\mathcal{F}^{geo}$ and $\mathcal{F}^{sem}$, are unprojected into 3D voxel space using Feature Line of Sight Projection ($\mathbf{FLoSP}$) \cite{cao2022monoscene}:
\begin{equation}
    \mathcal{F}^{geo, 3d} = \mathbf{FLoSP}(\mathcal{F}^{geo}),
\end{equation}
\begin{equation}
    \mathcal{F}^{sem, 3d} = \mathbf{FLoSP(\mathcal{F}}^{sem}), 
\end{equation}
 $\mathcal{F}^{sem, 3d} = \{ \mathcal{F}^{sem, 3d}_l, \in R^{N_l \times C \times H^{3d} \times W^{3d} \times L^{3d}} | l \in N_l \}$, $\mathcal{F}^{geo, 3d} = \{ \mathcal{F}^{geo, 3d}_l, \in R^{N_l \times C \times H^{3d} \times W^{3d} \times L^{3d}} | l \in N_l \}$.
The resulting 3D features are then processed by the proposed Frustum Mamba Decoder (\(\mathbf{FM_{Dec}}\)), which effectively captures long-range dependencies and refines the final 3D representation by leveraging structured spatial relationships within the frustum space. In \(\mathbf{FM_{Dec}}\), $\mathcal{F}^{sem, 3d}$ and $\mathcal{F}^{geo, 3d}$ are first fused into a unified 3D feature volume $\mathcal{F}^{3d}$. This volume is further refined by the Frustum Mamba Layer, which enhances spatial reasoning and improves the quality of the 3D representation by modeling long-range dependencies more effectively.

\textbf{Multi-scale information fusion.}
To achieve effective multi-scale feature fusion, we incorporate a channel attention mechanism that adaptively reweights features at different scales, enhancing their contribution based on contextual relevance. For $\mathcal{F}^{geo, 3d}$, we first apply global average pooling ($\mathbf{AvgPool}$) over spatial dimensions to obtain a compact representation \( \mathcal{F}^{geo, 3d}_{\text{avg}} \in \mathbb{R}^{N_l \times C} \), which is then mapped into weights $W^{geo} \in R^{N_l}$ with a Multilayer perception ($\mathbf{MLP}$):

\begin{equation}
\mathcal{F}^{geo, 3d}_{\text{avg}} = \mathbf{AvgPool}(\mathcal{F}^{geo, 3d}).
\end{equation}
\begin{equation}
    W^{geo} = \mathbf{MLP}(\mathcal{F}^{geo, 3d}_{\text{avg}})
\end{equation}
Finally, the softmax function is applied, and the output weights are used to compute a weighted average of multi-scale 3D feature volumes:
\begin{equation}
\mathcal{F}^{geo, 3d} = \sum_{n=1}^{N_l} \mathcal{F}^{geo, 3d}_l \cdot w_l, \quad w_l \in W^{geo}.
\end{equation}

$\mathcal{F}^{sem, 3d}$ will be gotten in a similar way.

\textbf{Multi-modality information fusion.}
The resulting two feature volumes corresponding to semantic and geometry information will be fused by elementwise addition:
\begin{equation}
  \mathcal{F}^{3d} = \mathcal{F}^{sem, 3d}  + \mathcal{F}^{geo, 3d}
\end{equation}

To mitigate the impact of 3D-to-2D projection ambiguity, which leads to information loss along the depth direction, the 3D model should possess strong global modeling capabilities to effectively capture long-range dependencies, enabling it to seek a globally optimal solution. Unfortunately, the transformer architecture is not suitable for this scenario, as its computational complexity scales quadratically with sequence length. This becomes impractical for 3D applications, where the large number of voxels leads to prohibitive memory and computational costs. Instead, we introduce the State Space Model (SSM) as an alternative solution to capture long-range dependencies in 3D space with linear scalability. 

Unlike transformers, Mamba requires reorganizing voxel data into a sequence based on their 3D coordinates before processing them sequentially. As shown in Fig. \ref{fig:discontinued}, this strategy introduces discontinuities due to the 2D-to-3D unprojection, which may lead to a performance drop.
To mitigate this issue, we propose a Frustum Reordering Strategy that reorganizes voxel data in frustum space.  
This is motivated by the fact that 3D voxel features are unprojected from 2D image features along rays, meaning they tend to be more structured along the image plane while exhibiting higher similarity along the depth direction.

$\mathbf{FM_{Dec}}$  adopts a 3D U-Net architecture following an Encoder-Decoder design. The proposed Frustum Mamba Layer, which combines the Frustum Reordering Strategy with a standard Mamba block, is applied in the encoder to enhance long-range spatial modeling. To ensure each voxel retains awareness of its 3D spatial position, a 3D convolution block is applied at the beginning to inject inductive bias. Specifically, the Frustum Mamba Layer can be formulated as:
\begin{equation}
    \mathcal{FF}^{3d} = \mathbf{Flatten}(\mathcal{F}^{3d})
\end{equation}
$\mathbf{Flatten}$ is the flatten operation and convert the input 3D feature volume $\mathcal{F}^{3d} \in \mathbb{R}^{C \times H^{3d} \times W^{3d} \times L^{3d}}$ into a voxel list $\mathcal{FF}^{3d} \in \mathbb{R}^{C \times (H^{3d} \times W^{3d} \times L^{3d})}$.
\begin{equation}
    \mathcal{SFF}^{3d} = \mathbf{FReorder}(\mathcal{FF}_i^{3d})
\end{equation}
$\mathbf{FReorder}$ represents the proposed Frustum Reorder Strategy, and the voxel sequence output from it can be represented as $\mathcal{SFF}^{3d}$, which will be processed by Mamba layer:
\begin{equation}
    \mathcal{SFF'}^{3d} = \mathbf{MambaLayers}(\mathcal{SFF}^{3d})
\end{equation}
\begin{equation}
    \mathcal{F'}^{3d} = \mathbf{Composit}(\mathcal{SFF'}^{3d})
\end{equation}
 $\mathbf{MambaLayers}$ denotes several standard Mamba layers, and $\mathbf{Composit}$ reconstructs the 3D feature volume $\mathcal{F'}^{3d}$ from the voxel list $\mathcal{SFF'}^{3d}$. Finally, the layer normalization $\mathbf{LN}$ and skip link will be added:
 \begin{equation}
    {\mathcal{F}'}^{3d} = \mathbf{LN}({\mathcal{F}}'^{3d}) + {\mathcal{F}}^{3d},
\end{equation}
After each Frustum Mamba Layer, a downsampling operation will be applied to enable $\mathbf{FM_{Dec}}$ encoder to extract more representative multi-scale 3D features.


The multi-scale 3D feature volumes from $\mathbf{FM_{Dec}}$'s encoder are fed into $\mathbf{FM_{Dec}}$'s decoder to restore the original volume size. Similar to the encoder, the decoder consists of multiple stages, each containing an upsampling block, 3D residual block and connected to its corresponding encoder block via skip links. To further enhance contextual information modeling, we incorporate the 3D Context Relation Prior (3D CRP) layer from \cite{cao2022monoscene} between the encoder and decoder. For additional implementation details, please refer to the supplemental material. Finally, the output feature volume is processed by a prediction head, which generates the final predictions.

\subsection{Supervision}
\label{sec:supervision}

The predicted occupancy from \(\mathbf{DM_{enc}}\) is supervised using the ground truth occupancy with the Binary Cross-Entropy (BCE) loss function $\mathcal{L}_{BCE}$. Additionally, following previous work, we apply cross-entropy loss $\mathcal{L}_{CE}$ for the final output, along with auxiliary supervision, including Scene-Class Affinity Loss $\mathcal{L}_{\text{scal}}^{\text{sem}}, \mathcal{L}_{\text{scal}}^{\text{geo}}$, Frustum Proportion Loss $\mathcal{L}_{\text{fp}}$ and Context Relation Loss $\mathcal{L}_{\text{rel}}$:
\begin{equation}
\mathcal{L}_{\text{total}} = \mathcal{L}_{\text{CE}} + \mathcal{L}_{\text{BCE}} + \mathcal{L}_{\text{rel}} + \mathcal{L}_{\text{scal}}^{\text{sem}} + \mathcal{L}_{\text{scal}}^{\text{geo}} + \mathcal{L}_{\text{fp}}.
\end{equation}
More details can be found in the supplemental materials.


\section{Experiments}
\label{sec:exp}

\begin{table}[!t]
    \centering
    \resizebox{\linewidth}{!}{
    \begin{tabular}{c|ccccc}
        Dataset & Ori-Res & Res & Classes & train / test\\
        \hline
        NYUv2 & 240$\times$144$\times$240 & 60$\times$36$\times$60 & 13 & 795 / 654 \\
        OSnet & 6$\times$60$\times$36 & 60$\times$60$\times$36 & 12 & 45755 / 19764\\
        OSnet-mini & 60$\times$60$\times$36 & 60$\times$60$\times$36 & 12 & 4639 / 2007 \\
    \end{tabular}}
    \vspace{-2mm}
    \caption{Dataset Summary. OSnet denotes Occ-ScanNet dataset.}
    \vspace{-6mm}
    \label{tab:dataset}
\end{table}

\begin{table*}[!t]
\centering
\resizebox{\linewidth}{!}{
\begin{tabular}{l|c|c|c c c c c c c c c c c|c}
\hline
\textbf{Method} & \textbf{Input} & \textbf{IoU} & \rotatebox{90}{\textcolor{brown}{\textbf{ceiling}}} & \rotatebox{90}{\textcolor{green}{\textbf{floor}}} & \rotatebox{90}{\textcolor{gray}{\textbf{wall}}} & \rotatebox{90}{\textcolor{cyan}{\textbf{window}}} & \rotatebox{90}{\textcolor{orange}{\textbf{chair}}} & \rotatebox{90}{\textcolor{red}{\textbf{bed}}} & \rotatebox{90}{\textcolor{purple}{\textbf{sofa}}} & \rotatebox{90}{\textcolor{yellow}{\textbf{table}}} & \rotatebox{90}{\textcolor{blue}{\textbf{tvs}}} & \rotatebox{90}{\textcolor{orange}{\textbf{furniture}}} & \rotatebox{90}{\textcolor{violet}{\textbf{objects}}} & \textbf{mIoU} \\
\hline
LMSCNet\textsuperscript{rgb} \cite{roldao2020lmscnet} & $\hat{x}^{occ}$ & 33.93 & 4.49 & 88.41 & 4.63 & 0.25 & 3.94 & 32.03 & 15.44 & 6.57 & 0.02 & 14.51 & 4.39 & 15.88 \\
AICNet\textsuperscript{rgb} \cite{li2020anisotropic} & $x_{rgb}, \hat{x}^{depth}$ & 30.03 & 7.58 & 82.97 & 9.15 & 0.05 & 6.93 & 35.87 & 22.94 & 11.10 & 0.71 & 15.90 & 6.45 & 18.15 \\
3DSketch\textsuperscript{rgb}\cite{chen20203d} & $x_{rgb}, \hat{x}^{TSDF}$ & 38.64 & 8.53 & 90.54 & 9.94 & 6.75 & 14.62 & 49.09 & 17.73 & 13.35 & 8.13 & 24.68 & 8.49 & 22.91 \\
MonoScene \cite{cao2022monoscene} & $x_{rgb}$ & 42.51 & 8.89 & 93.50 & 12.06 & 12.57 & 17.32 & 48.29 & 18.45 & 15.13 & 15.22 & 27.96 & 12.94 & 26.94 \\
NDC-Scene \cite{yao2023ndc} & $x_{rgb}$ & 44.17 & 12.02 & 93.51 & 13.11 & 13.77 & 15.83 & 49.57 & 39.87 & 17.17 & 24.57 & 31.00 & 14.96 & 29.03 \\
ISO\cite{yu2024monocular} & $x_{rgb}$ & 47.11 & \textbf{14.21} & \textbf{93.47} & \textbf{15.89} & 15.14 & 18.55 & 50.01 & 40.82 & 18.25 & \textbf{25.9} & 34.08 & \textbf{17.67} & 31.25 \\
\hline
GA-MonoSSC & $x_{rgb}$ & \textbf{47.51} & 12.15 & 93.39 & 15.73 & \textbf{17.1} & \textbf{19.23} & \textbf{52.16} & \textbf{44.64} & \textbf{20.63} & 25.55 & \textbf{35.22} & \textbf{19.71} & \textbf{32.32}
\end{tabular}}
\vspace{-2mm}
\caption{Quantitative results on the NYUv2 dataset.}
\vspace{-2mm}
\label{tab:nyu}
\end{table*}

\begin{table*}[!t]
\centering
\resizebox{\linewidth}{!}{
\begin{tabular}{l|c|c|c c c c c c c c c c c|c}
\hline
\textbf{Method} & \textbf{Input} & \textbf{IoU} & 
\rotatebox{90}{\textcolor{red}{\textbf{ceiling}}} & 
\rotatebox{90}{\textcolor{green}{\textbf{floor}}} & 
\rotatebox{90}{\textcolor{gray}{\textbf{wall}}} & 
\rotatebox{90}{\textcolor{cyan}{\textbf{window}}} & 
\rotatebox{90}{\textcolor{orange}{\textbf{chair}}} & 
\rotatebox{90}{\textcolor{red}{\textbf{bed}}} & 
\rotatebox{90}{\textcolor{purple}{\textbf{sofa}}} & 
\rotatebox{90}{\textcolor{yellow}{\textbf{table}}} & 
\rotatebox{90}{\textcolor{blue}{\textbf{tvs}}} & 
\rotatebox{90}{\textcolor{orange}{\textbf{furniture}}} & 
\rotatebox{90}{\textcolor{violet}{\textbf{objects}}} & 
\textbf{mIoU} \\
\hline
MonoScene \cite{cao2022monoscene} & $x_{rgb}$ & 41.60 & 15.17 & 44.71 & 22.41 & 12.55 & 26.11 & 27.03 & 35.91 & 28.32 & 6.57 & 32.16 & 19.84 & 24.62 \\
ISO\cite{yu2024monocular} & $x_{rgb}$ & 42.16 & 19.88 & 41.88 & 22.37 & 16.98 & 29.09 & 42.43 & 42.00 & 29.60 & 10.62 & 36.36 & 24.61 & 28.71\\
\hline
GA-MonoSSC & $x_{rgb}$ & \textbf{48.59} & \textbf{26.01} & \textbf{51.89} & \textbf{30.35} & \textbf{24.58} & \textbf{30.34} & \textbf{48.44} & \textbf{49.92} & \textbf{37.59} & \textbf{22.48} & \textbf{43.13} & \textbf{27.41} & \textbf{35.65} \\
\end{tabular}}
\vspace{-2mm}
\caption{Quantitative results on the Occ-ScanNet dataset.}
\vspace{-2mm}
\label{tab:occ}
\end{table*}

\textbf{Datasets.} We evaluate the proposed method on three widely used datasets for indoor monocular semantic scene completion: NYUv2 \cite{silberman2012indoor}, Occ-ScanNet \cite{yu2024monocular}, and Occ-ScanNet-mini \cite{yu2024monocular}. Detailed statistics for each dataset are presented in Tab. \ref{tab:dataset}. Notably, Occ-ScanNet-mini is a subset of Occ-ScanNet, and both adopt the original resolution for training and evaluation. For comparison, although a higher resolution is provided ($240\times144\times240$) for the NYUv2 dataset, the model is trained on a downsampled scene ($60\times36\times60$) to optimize computational efficiency while preserving essential scene details.

\textbf{Evaluation Metrics.} Consistent with standard practices, we evaluate performance on the Scene Completion (SC) task using Intersection over Union (IoU) for occupied voxels, irrespective of semantic class. For the Semantic Scene Completion (SSC) task, we report the mean IoU (mIoU) across all semantic classes.

\subsection{Comparison to state-of-the-art}

This section compares the proposed \textbf{GA-MonoSSC} with previous methods on the NYUv2 and Occ-ScanNet datasets. The experimental results are presented in Tab. \ref{tab:nyu} and Tab. \ref{tab:occ}, respectively. From Tab. \ref{tab:nyu}, We observe that the proposed MambaSSC achieves state-of-the-art performance in both IoU and mIoU, demonstrating its superior capability in recovering geometric structures while accurately predicting relevant semantic information. Notably, methods explicitly involving 3D data inputs (such as occupancy grids, depth maps, or Truncated Signed Distance Fields (TSDF)) demonstrate comparatively weaker performance. We attribute this limitation to the reduced capacity of their architectures, which may lack the robustness needed to capture the complex spatial relationships in 3D data effectively. We observe a trend of improved overall performance for methods that utilize monocular images as input, likely due to the adoption of more advanced architectures. Compared to the previous state-of-the-art method ISO, the proposed MambaSSC achieves a 3\% improvement in mIoU (31.25 vs 32.32) and a higher IoU (47.11 vs 47.51), demonstrating its effectiveness in MonoSSC. Furthermore, we observe that this improvement primarily stems from object classes. Specifically, for stuff classes (ceiling, floor, wall), ISO achieves slightly better overall performance. We assume this is because these classes typically have simple structures (e.g., flat planes), which are well-suited for CNN architectures that excel at local area modeling. In comparison, the proposed GA-MonoSSC outperforms ISO across almost all other object classes. This demonstrates that GA-MonoSSC is more effective in recovering and detecting nuanced geometric details and semantic information, thanks to its global awareness ability, which enables it to capture richer contextual information.

For the Occ-ScanNet dataset, we observe a similar trend but with a significantly larger performance gap. Specifically, GA-MonoSSC outperforms ISO by approximately 15\% in IoU (42.16 vs 48.59) and 24\% in mIoU (28.71 vs 35.65). The main reason for the significant performance difference between the NYU and Occ-ScanNet datasets lies in the dataset scale. The Occ-ScanNet dataset contains nearly 60× more samples than the NYU dataset, providing a substantially larger training set for learning richer representations. Furthermore, by equipping GA-MonoSSC with global context modeling in both the 2D and 3D domains through the Transformer and Mamba architectures, the model can efficiently scale to large-scale datasets. In contrast, CNN-based architectures, with their limited local receptive fields and fixed-weight encoding strategies, struggle to scale effectively, constraining their performance on larger datasets.

\begin{figure*}[t]
    \centering
    \includegraphics[width=\linewidth]{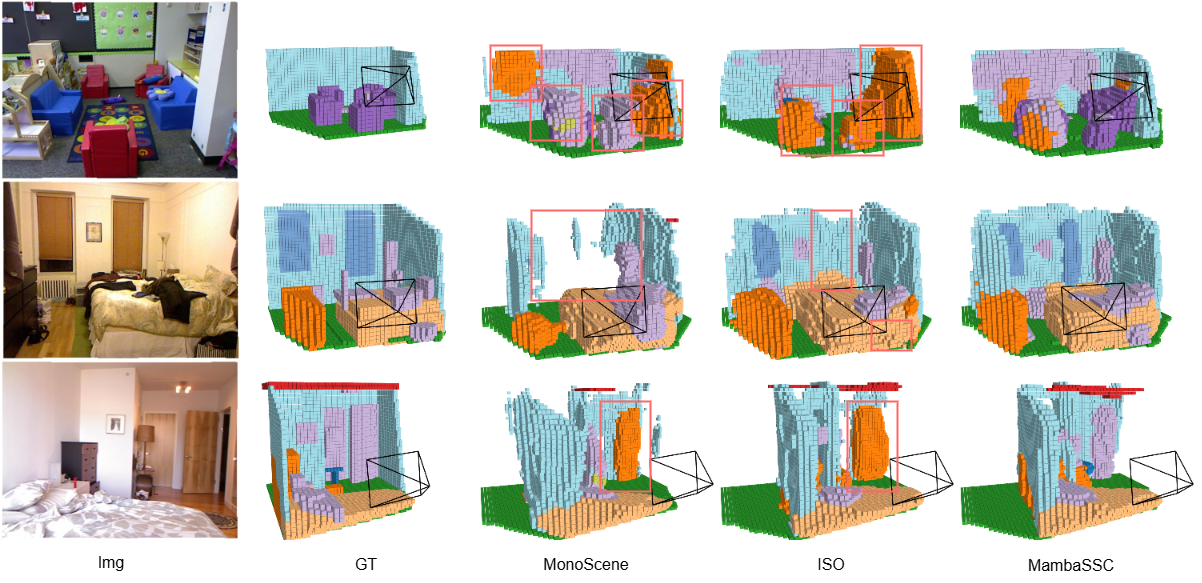}
    \vspace{-3mm}
    \caption{Qualitative results on NYU dataset. The \textcolor{red}{red} rectangle indicates incorrect semantic predictions. Compared to previous methods, the proposed MambaSSC produces significantly fewer incorrect predictions.}
    \vspace{-6mm}
    \label{fig:nyu_vis}
\end{figure*}

Finally, qualitative visualizations are presented in Fig. \ref{fig:nyu_vis} for the NYUv2 dataset. These visualizations demonstrate that the proposed GA-MonoSSC produces more accurate predictions. More qualitative results can be found in the supplemental materials.

\subsection{Ablation Study}
In this section, we evaluate the effectiveness of the proposed designs for both $\mathbf{DM_{Enc}}$ and $\mathbf{FM_{Dec}}$. 

\subsubsection{Effectiveness of DM$_\text{\textbf{Enc}}$}

\begin{table}[t]
\centering
\resizebox{\linewidth}{!}{
\begin{tabular}{c|c|cccc}
\hline
NYU & 2D Sup& N.A.& sem$\rightarrow$geo & geo$\rightarrow$sem & geo$\leftrightarrow$sem \\
\hline
mIoU  & 31.65 &  32.0 & 31.61 & 32.32 & 32.07 \\
IoU & 46.91 & 46.92 & 47.39 & 47.51 & 47.21 \\
\hline
\end{tabular}}
\caption{Ablation Study of $\mathbf{DM_{Enc}}$ on NYU dataset ``2D Sup'' means enforce $\mathbf{Dec_{geo}}$ to predict 2D depth instead 3D occupancy where the supervision is employed. ``N.A.'' means no interaction between $\mathbf{Dec_{geo}}$ and $\mathbf{Dec_{sem}}$. ``\textbf{A}$\rightarrow$\textbf{B}'' denotes the information is injected from \textbf{A} to \textbf{B}.}
\vspace{-6mm}
\label{tab:ablation_2d}
\end{table}

We first evaluate the effectiveness of the design of $\mathbf{DM_{Enc}}$  on the NYUv2 dataset, and the experimental results are presented in Tab. \ref{tab:ablation_2d}.
In this experiment, we first investigate the necessity of applying 3D supervision (occupancy) to the $\mathbf{Dec_{geo}}$ output rather than 2D supervision (depth). To enforce $\mathbf{Dec_{geo}}$ to explicitly learn geometry-related information, a relevant auxiliary supervision signal should be applied to distinguish it from $\mathbf{Dec_{sem}}$. Since $\mathbf{Dec_{geo}}$ produces a 2D feature map, a straightforward approach is to align it with the ground-truth 2D depth. However, as indicated by experiment result (2D Sup), this strategy does not achieve good performance. In contrast, converting predicted depth into 3D space (occupancy) and aligning it with  ground-truth occupancy yields better performance. We assume this is because the ultimate goal is to recover geometric structures in 3D space, and the latter approach enables $\mathbf{Dec_{geo}}$ to learn 3D-aware 2D features, whereas 2D supervision alone fails to provide sufficient 3D spatial awareness.

Next, we investigate the interaction between $\mathbf{Dec_{geo}}$ and $\mathbf{Dec_{sem}}$. We observe that the proposed method already achieves good performance even without interaction between $\mathbf{Dec_{geo}}$ and $\mathbf{Dec_{sem}}$ (N.A.). Injecting information from $\mathbf{Dec_{sem}}$ into $\mathbf{Dec_{geo}}$ leads to a noticeable increase in IoU, albeit at the cost of a performance drop in mIoU. This is reasonable, as $\mathbf{Dec_{geo}}$ benefits from high-level semantic information, which aids in recovering accurate geometric structures. However, without explicit geometry information,  semantic-related metric (mIoU) decreases accordingly. Instead, injecting geometry information into $\mathbf{Dec_{sem}}$ benefits both geometric and semantic performance. This indicates that enhancing semantic-related information contributes to an overall improvement in geometry performance as well. Finally, enabling $\mathbf{Dec_{geo}}$ and $\mathbf{Dec_{sem}}$  to interact deeply does not result in improved performance. We hypothesize that a strong interaction between  two modalities may lead to a performance drop, similar to previous approaches that jointly learn them. This is also evidenced by the poor performance observed in Tab. \ref{tab:ablation_3d}, Row 4.

\subsubsection{Effectiveness of FM$_\text{\textbf{Dec}}$}
Then, we evaluate the effectiveness of the design of $\mathbf{FM_{Dec}}$  on the NYUv2 and Occ-ScanNet-mini datasets, and the experimental results are presented in Tab. \ref{tab:ablation_3d}.

We first verify the necessity of utilizing the Mamba architecture, which effectively and efficiently captures long-range dependencies in 3D space. Its effectiveness is assessed by replacing all Mamba blocks with 3D convolution-based blocks, widely adopted in previous methods \cite{cao2022monoscene,yu2024monocular,yao2023ndc}, which consist of several bottleneck blocks. We observe that this replacement leads to a significant performance drop (Row 1 vs. Row 6), demonstrating the necessity of capturing global modeling ability in 3D space. 

Next, we evaluate the effectiveness of each component in $\mathbf{FM_{Dec}}$. We begin by verifying the necessity of multi-modality and multi-scale feature learning. We observe that on the Occ-ScanNet-mini dataset, omitting multi-scale features while jointly learning multi-modality features (instead of learning them separately) leads to a significant performance drop (Row 2 vs. Row 6). Learning multi-scale features primarily improves IoU (Row 2 vs. Row 4), likely because it enables the extraction of more detailed geometric features at different scales. Meanwhile, learning multi-modality features separately enhances both mIoU and IoU (Row 2 vs. Row 3), demonstrating that separately extracting these features allows the model to learn more nuanced information. Learning multi-scale features while separately learning multi-modality information yields the best overall performance, as shown in Row 6.
Similar conclusion can be drawn on NYU dataset, whereas the performance improvement is even larger. We assume that this is due to the limited amount of data in the NYU dataset, and that both strategies contribute to learning meaningful information in this scenario.
Finally, we highlight the importance of reordering voxels as sequential data in frustum space rather than in the original 3D space. We observe that applying the Mamba block to voxel sequences reordered in frustum space achieves better performance across all metrics on both datasets (Row 5 vs. Row 6). This demonstrates that ensuring feature continuity in the resulting voxel sequence, as achieved by the Frustum Reorder Strategy, is crucial for Mamba, which processes sequential data from start to end.

\begin{table}[!t]
\centering
\resizebox{\linewidth}{!}{
\begin{tabular}{c|c|ccc|cc|cc}
\hline
\multirow{2}{*}{\textbf{Row}} & \multirow{2}{*}{\textbf{Arch}} & \multirow{2}{*}{\textbf{MM}} & \multirow{2}{*}{\textbf{MS}} & \multirow{2}{*}{\textbf{FS}} & \multicolumn{2}{c|}{\textbf{NYUv2}} & \multicolumn{2}{c}{\textbf{OSnet-mini}} \\
\cline{6-9}
& & & & & \textbf{IoU} & \textbf{mIoU} & \textbf{IoU} & \textbf{mIoU} \\
\hline
0 & \textbf{ISO} & - & - & - & 47.11 & 31.25 & 51.03 & 39.08 \\
\hline
1 & \textbf{Conv} & \checkmark & \checkmark & - & 46.99 & 31.6 & 57.33 & 46.83 \\
\hline
2 & \multirow{5}{*}{\textbf{MB}} & & & \checkmark & 46.89 & 31.8 & 56.91 & 46.07 \\
3 & & \checkmark & & \checkmark & 47.02 & 32.0 & 58.42 & 47.39\\
4 & & & \checkmark & \checkmark & 46.96 & 31.81 & 57.59 & 46.2 \\
5 & & \checkmark & \checkmark &  & 47.1 & 31.9 & 57.51 & 46.66 \\
6 & & \checkmark & \checkmark & \checkmark &  47.51 & 32.32 & 58.97 & 48.19\\
\hline
\end{tabular}}
\caption{Ablation Study of $\mathbf{FM_{Dec}}$ on NYU and Occ-ScanNet-mini (\textbf{OSnet-mini}) datasets. \textbf{MB} denotes Mamba Block while \textbf{Conv} means replace Mamba Block with convolution block. \textbf{FS} denotes Frustum Reordering Strategy, \textbf{MM} denotes learning multi-modality features separately, \textbf{MS} denotes learning multi-scale features. We present ISO results as a reference, which show significantly lower performance compared to the proposed method.}
\vspace{-8mm}
\label{tab:ablation_3d}
\end{table}

\section{Conclusion}
\label{sec:conclusion}
In this work, we present GA-MonoSSC, a hybrid architecture for the MonoSSC task. Compared to previous CNN-based methods, which are limited by their local receptive fields and fixed-weight encoding strategies, GA-MonoSSC effectively captures global context in both the 2D image domain and 3D space. Specifically, it employs a Dual-Head Multi-Modality Encoder, based on a Transformer architecture, to capture spatial relationships among all features in the 2D image domain. The Frustum Mamba Decoder then processes 3D features using the State Space Model (SSM), which enables the efficient capture of long-range dependencies with linear computational complexity. Furthermore, we introduce a frustum reordering strategy within the Frustum Mamba Decoder to reduce feature discontinuities in the reordered voxel sequence, making it more compatible with the scan mechanism of the State Space Model (SSM). We conduct extensive experiments on Occ-ScanNet and NYUv2 datasets, achieving state-of-the-art performance, which validates the effectiveness of our design.


{
    \small
    \bibliographystyle{ieeenat_fullname}
    \bibliography{main}
}

\clearpage

\end{document}